\documentclass[
]{ceurart}

\usepackage{listings}
\usepackage{svg}
\usepackage{qrcode}
\begin{document}

\copyrightyear{2024}
\copyrightclause{Copyright for this paper by its authors.
  Use permitted under Creative Commons License Attribution 4.0
  International (CC BY 4.0).}

\conference{The 23rd International Semantic Web Conference, November 11--15, 2024, Hanover, MD}

\title{Here's Charlie! Realising the Semantic Web vision of Agents in the age of LLMs}

\author[1]{Wright, Jesse}[%
orcid=0000-0002-5771-988X,
email=jesse.wright@cs.ox.ac.uk,
url=https://www.cs.ox.ac.uk/people/jesse.wright/,
]
\address[1]{Computer Science Department, University of Oxford, UK}

\begin{abstract}
This paper presents our research towards a near-term future in which \textit{legal entities}, such as \textit{individuals} and \textit{organisations} can entrust semi-autonomous AI-driven agents to carry out online interactions on their behalf. The author’s research concerns the development of semi-autonomous Web agents, which consult users if and only if the system does not have sufficient context or confidence to proceed working autonomously. This creates a user-agent dialogue that allows the user to teach the agent about the information sources they trust, their data-sharing preferences, and their decision-making preferences. Ultimately, this enables the user to maximise control over their data and decisions while retaining the convenience of using agents, including those driven by LLMs.

In view of developing near-term solutions, the research seeks to answer the question: “How do we build a trustworthy and reliable network of semi-autonomous agents which represent individuals and organisations on the Web?”. After identifying key requirements, the paper presents a demo for a sample use case of a generic personal assistant. This is implemented using (Notation3) rules to enforce safety guarantees around \textit{belief}, \textit{data sharing} and \textit{data usage} and LLMs to allow natural language interaction with users and \textit{serendipitous} dialogues between software agents.
\end{abstract}


\maketitle

\section{Introduction}
There exists a substantial body of research on communication protocols for multi-agent systems, and it is reflected in the vision of the Semantic Web itself~\citep{lassila2001semantic, luke1997ontology, poslad2007specifying}
as shown by \href{https://www.w3.org/DesignIssues/Charlie.html}{Charlie}, the ``AI that works for you''. Yet, the 2006 lamentation that ``[b]ecause we haven’t yet delivered large-scale, agent-based mediation, some commentators argue that the Semantic Web has failed''~\citep{1637364} still rings true today. The growing use of LLMs raises a key challenge in building Trustworthy and Reliable Web Agents~\citep{deng2024large, sun2024trustllm}.  This is heightened by growing interest among LLM researchers in building dialogues between multiple LLMs~\citep{wu2023autogen, deng2023plug}. Moreover, recent research indicates the strong potential of the \href{https://blog.jeswr.org/2024/04/18/better-ai}{Semantic Web to complement emerging LLM technologies}~\citep{wright_old_2024}. For example, the use of Retrieval Augmented Generation (RAG) with Knowledge Graphs has shown to be effective in grounding LLM queries~\citep{kang2023knowledge}. The universal semantics and proof mechanisms of the Semantic Web stack are therefore pertinent to the successful development of semi-autonomous Web agents using LLMs. 

\section{Design Requirements}~\label{sec:design_requirements}
We identify the following non-functional requirements for an agent communication protocol. It must be possible for semi-autonomous agents to:
\begin{enumerate}
\item \textit{Identify} legal entities, such as individuals or organisations, on the Web~\citep{webid} so they can be referenced.
\item \textit{Deterministically discover} other agents representing an entity from their Web identity~\citep{webid}. This does \textit{not} require all agents to be publicly advertised; some may be discovered from links to protected documents.
\item Describe, and agree to, any \textit{usage controls}~\citep{acp, Iannella_Villata_2023, pandit2019creating} associated with data they exchange. This allows sharing of protected data while articulating the recipient's legal or moral obligations~\citep{wright2024wantcookieautomatedtransparent}.
\item Describe the \textit{origin} and \textit{provenance} of data they exchange. In an open world of agents that can ``say anything about anything,'' systems can identify which external claims to believe for a given task, based on the agent's internal trust model.
\item \textit{Unambiguously} describe \textit{ground truths} they send, and \textit{agreements} they make, using a formal representation. Consider the case where an individual's agent purchases a flight from an airline's agent. Structured ground truths eliminate an LLM's risk of hallucination or misinterpretation of key information, such as the flight time (``10 o'clock'' could be 22:00 or 10:00). As agents represent entities in binding agreements, this approach also reduces the risk of legal disputes by limiting the subjectivity of agreed terms and thus the ability to reinterpret or rescind them~\citep{forbesWhatCanada}. Furthermore, agents can implement rule-based internal safeguards, such as user-defined daily spending limits. Truly generic agents may generate and communicate structured ontologies when encountering new tasks. In many cases we expect LLM-supported ontology construction~\citep{kommineni2024humanexpertsmachinesllm} to facilitate generation; however, research is required to understand how (1) agents can align on conceptual models for use and (2) how human oversight can be maintained without disrupting user experience.

\item Contextualise a task which may be \textit{ambiguous} or poorly defined, such that interacting agents can introduce new solution spaces or negotiating actors in a \textit{serendipitous} manner. 
\end{enumerate}

\section{Sample Use-Case and Implementation}~\label{implementation}

\begin{wrapfigure}{r}{2cm} 
  \vspace{-12pt} 
  \qrcode{https://github.com/jeswr/phd-language-dialogue-experiment}
\end{wrapfigure}

We implemented the following flow where agents act as personal assistants for individual users: 
\begin{enumerate}
    \item Jun types into a chat ``Schedule a meeting with Nigel next week'';
    \item Jun's agent identifies data to be shared with Nigel and requests relevant sharing permissions from Jun (where not already obtained);
    \item Nigel's agent receives a request from Jun;
    \item Nigel is prompted to confirm that he believes Jun is an authoritative source of truth for her calendar (where not already obtained);
    \item Nigels agent proposes a meeting time to Nigel; and
    \item the meeting is proposed to Jun's agent and automatically confirmed.
\end{enumerate}

\begin{figure}
    \centering
    \includegraphics[width=\linewidth]{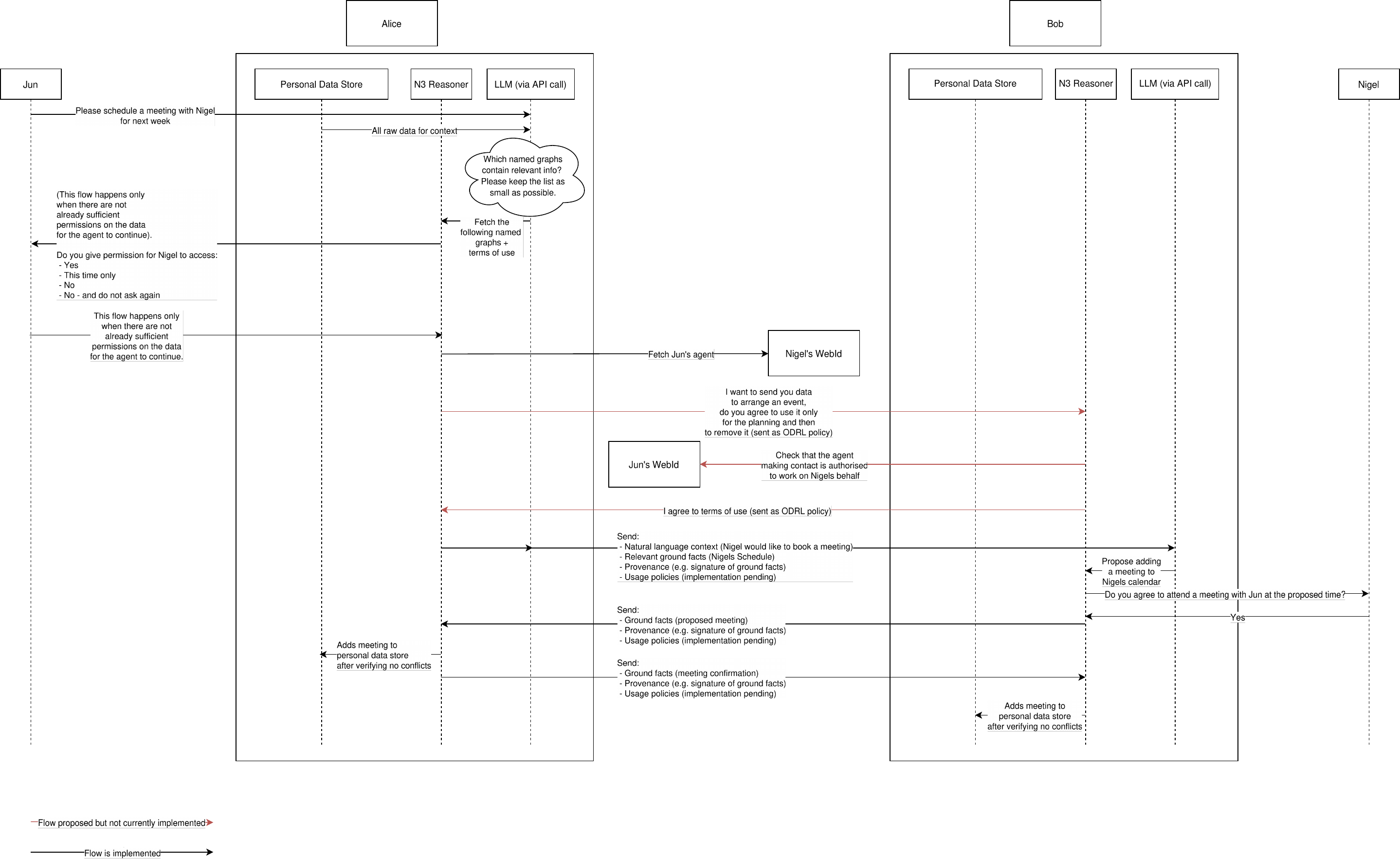}
    \caption{Flow diagram for the scheduling use case. Alice is Jun's agent and Bob is Nigel's agent.}
    \label{fig:flow}
\end{figure}

\newpage

We have created a running demo with a \href{https://github.com/jeswr/phd-language-dialogue-experiment/tree/main#demo-video}{video}, \href{https://raw.githubusercontent.com/jeswr/phd-language-dialogue-experiment/main/flow-diagram.svg}{flow-diagrams} (including Figure~\ref{fig:flow}) and \href{https://linktr.ee/semiautoweb}{other resources} for \href{https://github.com/jeswr/phd-language-dialogue-experiment}{our codebase}\footnote{\url{https://github.com/jeswr/phd-language-dialogue-experiment}}. The implementation corresponds to the above use-case steps:

\begin{enumerate}
\item Given the user prompt and a set of known WebID profiles~\citep{webid}, an LLM called by Jun's agent identifies the relevant entities for the agent to negotiate with (Nigel), and the WebIDs of those entities. Given the user prompt, and the user's personal knowledge graph, an LLM called by Jun's agent identifies which subset (as a list of named graphs) of the user data are needed to fulfil the user's request.

\item Notation3~\citep{berners1998notation3} reasoning is used to identify the policies applicable to the data subset. In the available demo recording, policies are encoded in ACP~\citep{acp}; we are currently migrating to use ODRL~\citep{Iannella_Villata_2023} and DPV~\citep{pandit2019creating}. If these policies do not yet permit read access to Nigel, Jun is prompted to modify them. Jun's agent then dereferences Nigel's WebID~\citep{webid} to discover information about his agent.

\item Jun's agent uses an LLM to construct a message for Nigel's agent, explaining the context of Jun's task: ``Jun seeks to schedule a meeting for next week. Propose a time for Jun and Nigel to meet using their calendars.'' Jun's agent sends Nigel's agent this message along with the RDF description of Jun's calendar and any associated policies and provenance. With ACL, Nigel's agent does not need to agree to any policy obligations; this changes with ODRL. The provenance in this case is simply a signature of the canonicalised calendar dataset~\citep{rdfcanon} using Jun's public key.

\item As Nigel has instructed his agent that Jun is an authoritative source of information on all topics, his agent \textit{believes} (takes as ground truth) the signed RDF dataset sent by her agent. We are developing conceptual models for agentic trust; these extend existing trust vocabularies~\citep{richardson2003trust, galizia2006wsto, sherchan2010trust, amaral2019towards} with a range of features including (1) qualifying whether sources are trusted for particular \textit{types} of claims; for instance, most agents should trust certified airlines to present flight times and prices, but not medical data (2) qualifying the forms of provenance \textit{secure} enough for a given task; for instance, an insurance provider may require provenance demonstrating a user was signed in with two-factor authentication when entering financial details to their knowledge base.

\item Nigel's agent proposes a meeting time, using the natural language context (\textit{not} a ground truth) and the calendar dataset (ground truth). The LLM proposes a meeting time, then the N3 reasoner applies rules to (1) ensure no calendar conflicts and (2) check for user confirmation, before adding the proposed time to the knowledge base. In a future iteration, we plan to use the LLM to generate an N3 query that proposes a meeting time based on Nigel's Personal Data Store and Jun's calendar.

\item Upon meeting the above requirements, the reasoner sends to Jun's agent a meeting proposal, in the form of an RDF dataset with attached usage policies and provenance. Jun's agent confirms this dataset can be believed based on the internal trust model. The rules within Jun's agent validate that there are no conflicting events. Jun's personal knowledge base is updated with the event, and a confirmation is sent to Nigel's agent.
\end{enumerate}

\section{Conclusion and Future Research}
We have implemented a generic personal assistant that communicates using a protocol satisfying the requirements of Section~\ref{sec:design_requirements}. Future work will make the design requirements more rigorous by (1) gathering requirements for personal agents through user studies, and (2) engaging with industry to develop specialised agents, including product sales agents. Concurrently, we shall formalise the vocabularies for exchanging \textit{provenance} and \textit{terms of use} between agents and modelling \textit{trust} and \textit{data policies} within agents, extending those vocabularies discussed in Section~\ref{implementation}. Once these vocabularies mature, we will develop reasoning specifications to mediate between the internal representations and exchanged metadata. This enables agents to negotiate to obtain sufficient provenance to believe claims, and find agreeable data terms of use between agents - whilst concurrently updating their internal models via user interaction.

\section*{Acknowledgements}

Jesse Wright is funded by the \href{https://www.cs.ox.ac.uk/}{Department of Computer Science}, \href{https://www.ox.ac.uk/}{University of Oxford}.

\newpage

\bibliography{sample-ceur}

\end{document}